\pdfoutput=1

\documentclass[11pt]{article}
\usepackage{graphicx, tabularx}

\usepackage[]{acl} 

\usepackage{times}
\usepackage{latexsym}

\usepackage{tabularx}
\usepackage{amsmath}
\usepackage{multirow,bigdelim}
\newcolumntype{R}{>{\raggedleft\arraybackslash}X}
\newcolumntype{Y}{>{\centering\arraybackslash}X}

\usepackage[T1]{fontenc}

\usepackage[utf8]{inputenc}

\usepackage{microtype}

\title{Enriching Abusive Language Detection with Community Context}

\author{
  Jana Kurrek \textsuperscript{\dag} \\
  McGill University \\
  School of Computer Science \\
  \texttt{\small jana.kurrek@mail.mcgill.ca} \\ \And
  Haji Mohammad Saleem \textsuperscript{\dag} \\
  McGill University \\
  School of Computer Science \\ 
  \texttt{\small haji.saleem@mail.mcgill.ca} \\ \And
  Derek Ruths \\
  McGill University \\
  School of Computer Science \\
  \texttt{\small derek.ruths@mcgill.ca} \\}

\newcommand\blfootnote[1]{%
  \begingroup
  \renewcommand\thefootnote{}\footnote{#1}%
  \addtocounter{footnote}{-1}%
  \endgroup
}

\begin{document}
\maketitle
\begin{abstract}
Uses \blfootnote{\textsuperscript{\dag} These authors made equal contributions.} of pejorative expressions can be benign or actively empowering. When models for abuse detection misclassify these expressions as derogatory, they inadvertently censor productive conversations held by marginalized groups. One way to engage with non-dominant perspectives is to add context around conversations. Previous research has leveraged user- and thread-level features, but it often neglects the spaces within which productive conversations take place. Our paper highlights how community context can improve classification outcomes in abusive language detection. We make two main contributions to this end. First, we demonstrate that online communities cluster by the nature of their support towards victims of abuse. Second, we establish how community context improves accuracy and reduces the false positive rates of state-of-the-art abusive language classifiers. These findings suggest a promising direction for context-aware models in abusive language research.
\end{abstract}

\section{Introduction}
Existing models for abuse detection struggle to grasp subtle knowledge about the social environments that they operate within. They do not perform natural language understanding and consequently cannot generalize when tested out-of-distribution \cite{bender2021dangers}. This problem is often the result of training data imbalance, which encourages language models to overestimate the significance of certain lexical cues. For instance, \citet{wiegand2019detection} observe that ``commentator'', ``football'', and ``announcer'' end up strongly correlated with hateful tweets in the \citet{waseem2016hateful} corpus. This trend is caused by focused sampling, and it does not reflect an underlying property of abusive expressions.

When models rely on pejorative or demographic words, they can encode systemic bias through \textit{false positives} \cite{kennedy2020contextualizing}. For example, research has established that detection algorithms are more likely to classify comments written in African-American Vernacular English (AAVE) as offensive \cite{davidson2019racial, xia2020demoting}. Benign tweets like ``\textit{Wussup, n*gga}!'' and ``\textit{I saw his ass yesterday}'' both score above 90\% for toxicity \cite{sap2019risk}. Similarly, \citet{zhang2020demographics} analyze the Wikipedia Talk Pages Corpus \cite{dixon2018measuring} and find that 58\% of comments that contain the term ``gay'' are labelled as toxic, while only 10\% of all comments are toxic. This enables the misclassification of positive phrases like ``\textit{she makes me happy to be gay}''. Even Twitter accounts belonging to drag queens have been rated higher in terms of average toxicity than the accounts associated with white nationalists \cite{oliva2021fighting}. These findings underline how language models with faulty correlations can facilitate the censorship of productive conversations held by marginalized communities.

Productive conversations containing slurs are common, and they take many forms \cite{hom2008semantics}. Research inspired by the \#MeToo movement has focused on the detection of sexual harassment disclosures by victims \cite{deal2020definitely}, but this research has not been sufficiently integrated into the literature on abusive language detection. The distinction between actual sexist messages and messages calling out sexism is rarely addressed in the field \cite{chiril2020he}. A similar trend is seen with sarcasm. Humor and self-irony can be employed as coping mechanisms by victims of abuse \cite{garrick2006humor}, yet they constitute frequent sources of error for state-of-the-art classifiers \cite{vidgen2019challenges}. For example, the median toxicity score for language on \texttt{transgendercirclejerk}, a ``parody [subreddit] for trans people'', is as high as 90\% \cite{kurrek2020towards}. More broadly, transgender users are ``excluded, harmed, and misrepresented in existing platforms, algorithms, and research methods'' related to network analysis \cite{stewart2021manosphere}.

Meaningful improvements in abusive language detection require a  thoughtful engagement with the perspectives of marginalized communities and their allies. One way to ensure that machine learning frameworks are socially conscientious is to add context around conversations. Past research has explored the contextual information within conversation threads \cite{pavlopoulos2020toxicity, ziems2020aggressive}, user demographics \cite{unsvaag2018effects, founta2019unified}, user history \cite{saveski2021structure, qian2018leveraging, dadvar2013improving}, user profiles \cite{unsvaag2018effects, founta2019unified}, and user networks \cite{ziems2020aggressive, mishra2018author} with varying degrees of success in improving performance. However, most modelling efforts for abusive language detection neglect one major aspect of online conversations: the community environment they take place within.

Online communities adhere to a variety of sociologistic norms that reinforce their identities. This phenomenon is easily observed on Reddit, where community structure is an explicit component of platform design. For example, the majority of comments on the pro-Trump subreddit \texttt{The\_Donald} delegitimize liberal ideas \cite{mclamore2020social, soliman2019characterization}. Similarly, a collection of ``manosphere'' subreddits espouse misogynistic ideologies  \cite{stewart2021manosphere, ging2019alphas}. More broadly, communities can reinforce ``toxic technocultures" \cite{massanari2017gamergate}, and those technocultures are not limited to Reddit. Community structure is present across 4chan, Facebook, Voat, etc., and it exists in a less explicit manner on platforms like Twitter \cite{silva2017methodology}.

In this paper, we study community context on Reddit, and we focus specifically on language that is collected using slurs. We demonstrate that subreddits cluster by the nature of their support towards marginalized groups, and we use subreddit embeddings to improve the accuracy and false positive rates of state-of-the-art abusive language classifiers. While our analysis is platform-specific, it suggests a promising new direction for context-aware models.

\begin{figure*}[t!]
 \centering
  \includegraphics[width=16cm]{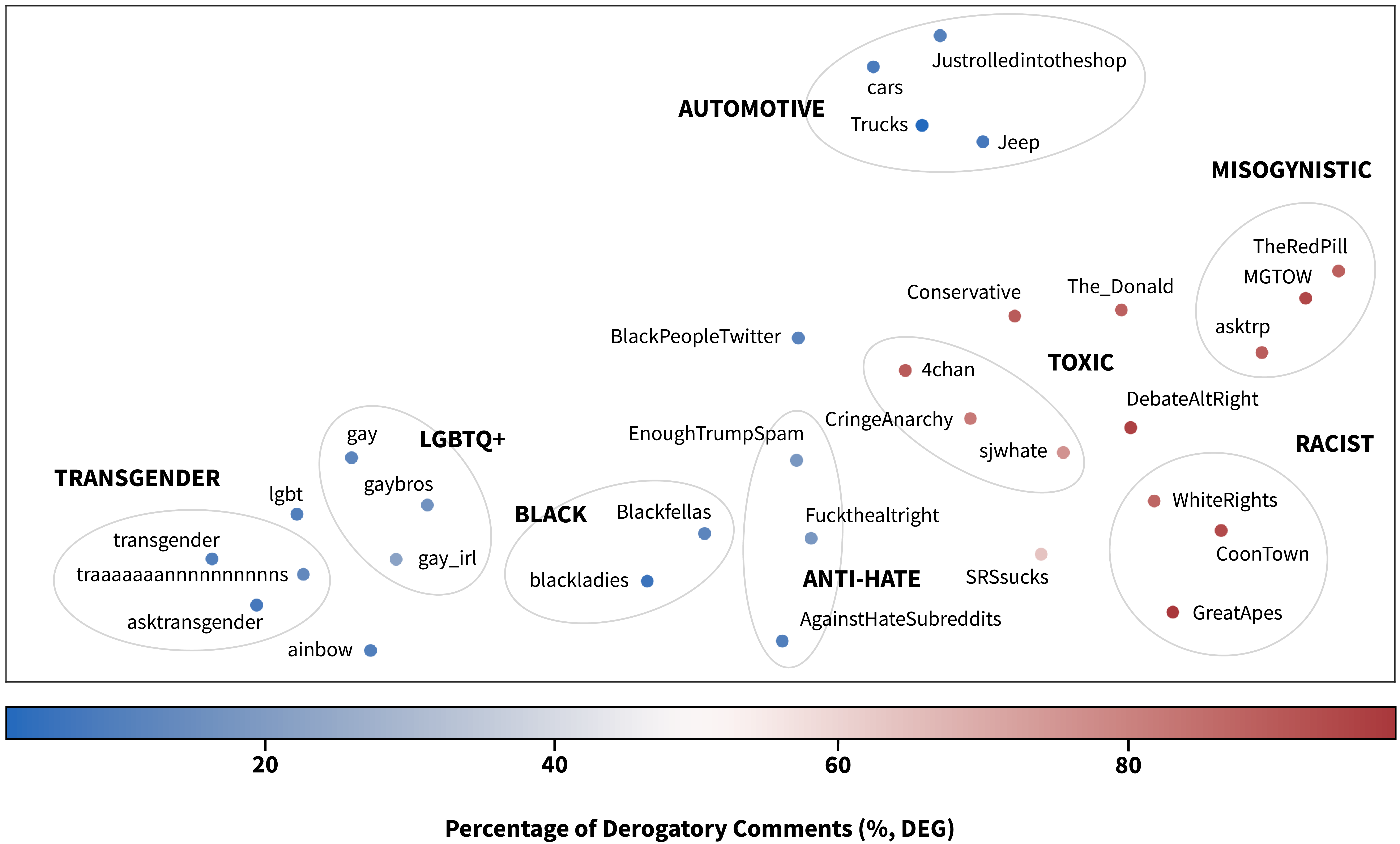}
    \caption{A subset of our subreddit embeddings plotted in two-dimensions using UMAP. Community clusters emerge based on the nature of users' support towards marginalized groups.}
\label{fig:sub-cluster}
\end{figure*}

\section{Related Work}
\subsection{Methods in Abusive Language Detection}
Abusive language detection is a relatively new field of research, with ``very limited'' work from as recently as 2016 \cite{waseem2016hateful}. Early methods featured Naive Bayes \cite{liu2014combining}, SVMs \cite{tulkens2016dictionary}, Random Forests \cite{warner2012detecting}, Decision Trees \cite{del2017hate}, and Logistic Regression \cite{burnap2014hate, greevy2004automatic}.

However, recent developments in NLP have directed the field towards neural and Transformer-based approaches. CNNs, LSTMs (+ Attention), and GRUs have been widely used in the literature \cite{mathur2018detecting, meyer2019platform, chakrabarty2019pay, zhang2018detecting, modha2018filtering}. As of 2019, researchers have begun adopting pre-trained language models. Contemporary work leverages BERT, DistilBERT, ALBERT, RoBERTA, and mBERT \cite{alonso2020hate, davidson2020developing}. In fact,  \citet{bodapati2019neural} note that seven of the top ten performing models for offensive language identification at SEMEVAL-2019 were BERT-based. A similar trend was seen at SEMEVAL-2020, where ``most teams used some kind of pre-trained Transformers'' \cite{zampieri-etal-2020-semeval}. Regardless of architecture, methods in abusive language detection can be divided into content- and context- based approaches.

Content-based approaches rely on comment text for feature engineering. Researchers have used TF-IDF weighted n-gram counts as well as distributional embeddings for text representation \cite{davidson2017automated, nobata2016abusive, van2018automatic}, POS tags or dependency relations for encoding syntactic information \cite{narang2020abusive}, and the frequencies of hashtags, URLs, user mentions, emojis, etc. for detecting platform-specific tokens. Lexicons are also popular for capturing sentiment, politeness, emotion, and hate words \cite{cao2020deephate, nobata2016abusive, markov2021improving, koufakou-etal-2020-hurtbert}. The central assumption behind content-based abusive language detection is that comments can be exclusively assessed using textual features. However, this assumption neither holds in theory nor in practice because linguistic structures are discourse-determined, and that discourse is shaped by social, historical, and political context \cite{bridges2017gendering}. Semantics cannot be completely interpreted using content cues alone. Even human annotators struggle to classify comments that involve satire or homonymy in the absence of broader context \cite{kurrek2020towards}. In light of these concerns, researchers are increasingly identifying the importance of user and conversational features to their detection frameworks. We summarize five main trends in the literature below.

\noindent \textbf{Conversational Context.} Attempts have been made to situate abusive comments within conversation threads. Threads have been studied using preceding comments \cite{pavlopoulos2020toxicity, karan2019preemptive}, discussion titles \cite{gao2017detecting}, and counts for aggressive comments \cite{ziems2020aggressive}. The position of a comment in a thread - start or end - has also been considered \cite{joksimovic2019automated}. Finally, researchers have analyzed conversation graphs for topological indicators of abuse \cite{papegnies2017graph}.

\noindent \textbf{User Demographics.} Researchers have attempted to incorporate user-level context through demographic signals for age, location, and gender. Age has been extracted from user disclosures, but these disclosures can be unreliable when users have an incentive to view adult-rated content \cite{dadvar2013improving}. Previous work has inferred gender from user names \cite{waseem2016hateful, unsvaag2018effects}, expressions in user biographies \cite{waseem2016hateful, unsvaag2018effects}, and in-game avatar choices \cite{balci2015automatic}, but these methods can fail when names are gender-neutral. Location information obtained through geo-coding has also been used to analyze hateful tweets \cite{fan2020stigmatization}.

\noindent \textbf{User History.} Patterns in user behaviour, including daily logins \cite{balci2015automatic}, favourites \cite{unsvaag2018effects}, and posting history \cite{saveski2021structure, ziems2020aggressive}, can be used as features in abusive language detection models. Some work focuses directly on the content of past comments. For example, \citet{dadvar2013improving} look for the prevalence of profanity in text. Conversely, \citet{qian2018leveraging} encode all historical posts by a user. Similarly, \citet{ziems2020aggressive} create TF-IDF vectors derived from a user's timeline.

\noindent \textbf{User Profiles.} Several elements of profile metadata have been studied as a proxy for digital identity. These elements include usernames \cite{gao2017detecting}, user anonymity, the presence of updated profile pictures \cite{unsvaag2018effects}, biographies \cite{miro2018hate}, verified account status \cite{ziems2020aggressive}, counts for followers \cite{founta2019unified}, and friends \cite{balci2015automatic}. Some other profile features include profile language \cite{galan2016supervised} and account age \cite{founta2019unified}.

\noindent \textbf{User Networks.} Homophily in social networks induces user clusters based on shared identities. These clusters have been shown to represent collective ideologies and moralities \cite{dehghani2016purity}, motivating researchers to examine local user networks for markers of abusive behaviour. Interaction and connection-based social graphs are analyzed using Jaccard's similarity and eigenvalue or closeness centrality \cite{ziems2020aggressive, chat2017mean, founta2019unified, unsvaag2018effects, papegnies2017graph}, which are also relevant for creating user embeddings.

\subsection{Methods in Community Profiling}
Network data may capture localized trends about individual users, but it often overlooks how groups of users behave as a whole. There are connection- and content-based solutions for explicit community profiling which, to the best of our knowledge, exist outside of contemporary abusive language research. Connection-based solutions evolved out of the idea that similar communities house similar users. In contrast, content-based solutions claim that similar communities contain similar content.

\noindent \textbf{Connection-based Representations.} Vector representations of online communities are known to encode semantics \cite{martin2017community2vec}. Popular techniques for obtaining these representations require the construction of a community graph. \citet{kumar2018community} construct a bipartite multigraph between Reddit users and subreddits. An edge $u_i \rightarrow s_j$ is added for each post by a user $u_i$ in a subreddit $s_j$. The graph is then used to learn subreddit embeddings by a ``node2vec-style'' approach.

\citet{martin2017community2vec} creates a symmetric matrix of subreddit-subreddit user co-occurrences, where $\textbf{X}_{ij}$ is the number of unique users who have commented at least ten times in the subreddits $i$ and $j$. Skip-grams with negative sampling or \texttt{GloVe}  can then be used to obtain subreddit embeddings. Here, subreddits and user co-occurrences inherit the role of words and word co-occurrences respectively. \citet{waller2019generalists} also treat communities as ``words'' and users who comment in them as ``contexts'' and adapt word2vec for community representations. The subreddit graph proposed in \citet{janchevski2019study} contains edges weighted by the number of shared users between the two subreddits. They only consider users who participate in at least ten subreddits and use node2vec to generate node embeddings.

\noindent \textbf{Content-based Representations.} Content-based solutions for community profiling rely on methods for document similarity. \citet{janchevski2019study} average the word2vec representations for the top 30 words in each subreddit, ranked by TF-IDF score. This research is currently limited, relative to other techniques.

\section{Methodology}

\subsection{Corpus}
We select the \texttt{Slur-Corpus} by \citet{kurrek2020towards}. It consists of ~40k human-annotated Reddit comments. Every comment contains a slur and is labelled as either derogatory (\texttt{DEG}), appropriative (\texttt{APR}), non-derogatory non-appropriative (\texttt{NDNA}), or homonym (\texttt{HOM}). The corpus is nearly evenly split between derogatory and non-derogatory (\texttt{APR, NDNA, HOM}) slur usages, with 51\% of comments labelled \texttt{DEG} (see Table \ref{tab:corpus_details}). 

The \texttt{Slur-Corpus} is one of few community-aware resources for abusive language detection. The data is sampled over the course of a decade (October 2007 to September 2019), reflecting a variety of users and language conventions. Every comment is published with the subreddit from which it was sourced, and the authors curate content across a number of antagonistic, supportive, and general discussion communities. As opposed to random sampling, this method guarantees the representation of targeted and minority voices. We see this as crucial for investigating the role of social context within abusive language conventions.

\subsection{Definitions}
Subreddits are niche communities dedicated to the discussion of a particular topic, with users participating in subreddits that engage their personal interests. As a result, subreddits often exhibit language specificity that can be leveraged for making inferences about slur usages.

Consider the slur \textit{tr*nny}. The comment, ``\textit{I am genuinly surpised at a suicidal tr*nny}'' from \texttt{CringeAnarchy} is derogatory. In contrast, ``\textit{So do I. Just that the tr*nny is dying on me lol.}'' from \texttt{Honda} is non-derogatory because \textit{tr*nny} is being used as a homonym. Both of these subreddits adhere to different linguistic norms and appeal to different user bases. Quantifying these differences is important. Niche or small automotive subreddits are likely to be related to \texttt{Honda}, and their users may also use \textit{tr*nny} to mean \textit{transmission}.

\subsection{Constructing Subreddit Embeddings}
\label{sec:glove}

We construct subreddit embeddings based on user comment co-occurrence. This method aligns with prior work on the subject \cite{martin2017community2vec, kumar2018community, waller2019generalists}, but extends it by considering data collected at a much larger scale. We use all publicly available Reddit comments prior to September 2019 in order to generate lists of users that comment in each found subreddit \cite{baumgartner2020pushshift}. We then store frequency counts for each list and, in total, identify 998K unique subreddits and 42.7M unique authors over the course of 12 years. There is a long tail because many subreddits have low participation.

Next, we identify active users, defined as being any users with at least ten comments in a subreddit. We exclude bot accounts and focus on top subreddits by activity. This leaves 10.4K subreddits and 12.2M unique users. With this data, we build a subreddit adjacency matrix $\textbf{A}$, where $\textbf{A}_{ij}$ is the number of co-occurring users in subreddits $i$ and $j$. We use \texttt{GloVe} to generate dense embeddings from $\textbf{A}$, and we run it over 100 epochs with a learning rate of 0.05 and a representation size of 150.

\subsection{Evaluating Subreddit Embeddings}

\begin{table}[t]
\begin{tabularx}{\linewidth}{XRc|XR}
\textbf{Label}           & \textbf{Count}    & \textbf{\%}               & \textbf{Stats}& \textbf{Count}    \\ \hline
\texttt{DEG}             & 20531             & 51\%                      & Users         & 36962             \\ \cline{1-3}
\texttt{NDNA}            & 16729             &                           & Posts         & 34610             \\
\texttt{HOM}             & 1998              & 49\%                      & Subreddits    & 2691              \\ 
\texttt{APR}             & 553               &                           &               &                   \\ \cline{1-3}
\textit{Total}           & 39811             &                           &               &
\end{tabularx}
\caption{Characteristics of the \texttt{Slur-Corpus}. The split between \texttt{DEG} and \texttt{NDG} comments is nearly equal.}
\label{tab:corpus_details}
\end{table}

Our tests for subreddit similarity seek to capture two conditions: (1) compositionality: similar subreddits have similar constituent subreddits; and (2) analogy: subreddit similarity is preserved under analogical argument. We rely on vector algebra to model each of these two conditions.

\subsubsection{Similarity} The similarity between subreddits $S_i$ and $S_j$ is simply the cosine similarity of their representations:

\[sim(S_i, S_j) = \frac{\vec{S_i}\cdot\vec{S_j}}{|\vec{S_i}||\vec{S_j}|}\]

\subsubsection{Composition Tests}
We find a subreddit $S_k$ that represents the sum of $S_i$ and $S_j$. We create $\vec{V} = \vec{S_i} + \vec{S_j}$, and then compute $S_k := max_x(\{sim(\vec{V}, \vec{S_x})\})$. We run the composition test to identify local sports team subreddits from combinations of sport and city subreddits ($\overrightarrow{sport} + \overrightarrow{city} = \overrightarrow{team}$). We base these tests on the evaluations of \citet{martin2017community2vec}.

\subsubsection{Analogy Tests}
We find a subreddit $S_n$ such that $\vec{S_i} : \vec{S_j} :: \vec{S_m} : \vec{S_n}$ for a triad of subreddits $S_i$, $S_j$ and $S_m$. We create $\vec{V} = \vec{S_i} - \vec{S_j} + \vec{S_m}$ and then compute $S_n = max_x(\{sim(\vec{V}, \vec{S_x})\})$. The analogy tests, based on \citet{waller2019generalists}, identify: 

\begin{enumerate}
    \item A local team given a city and sport:
    
    $\overrightarrow{city}: \overrightarrow{team} :: \overrightarrow{city'} : \overrightarrow{team'}$
    
    \item A sport given a team and its city:
    
    $\overrightarrow{team}: \overrightarrow{sport} :: \overrightarrow{team'} : \overrightarrow{sport'}$
    
    \item A city given a university
    
    $\overrightarrow{university}: \overrightarrow{city} :: \overrightarrow{university'} : \overrightarrow{city'}$
    
\end{enumerate}

\begin{table*}[th!]
\centering
\scriptsize
\begin{tabularx}{\linewidth}{rclcl|RcXcRcX}
\multicolumn{5}{c}{}                                 & \multicolumn{7}{c}{}                                                                           \\ \hline
\texttt{\textbf{city}}      & + & \texttt{\textbf{sport}}      & = & \texttt{\textbf{team}}              & \texttt{\textbf{city}}       & : & \texttt{\textbf{team}}            & :: & \texttt{\textbf{city}}            & : & \texttt{\textbf{team}}              \\ \hline

\texttt{toronto} & + & \texttt{baseball} & = & \texttt{Torontobluejays} & \texttt{boston}   & : & \texttt{BostonBruins}  & :: & \texttt{toronto}       & : & \texttt{leafs}           \\
\texttt{chicago} & + & \texttt{baseball} & = & \texttt{CHICubs}         & \texttt{boston}   & : & \texttt{Patriots}      & :: & \texttt{chicago}       & : & \texttt{CHIBears}        \\ \cline{6-12}
\texttt{chicago} & + & \texttt{hockey}   & = & \texttt{hawks}           & \texttt{\textbf{team}}       & : & \texttt{\textbf{sport}}           & :: & \texttt{\textbf{team}}            & : & \texttt{\textbf{sport}}             \\ \cline{6-12}
\texttt{chicago} & + & \texttt{nba}      & = & \texttt{chicagobulls}    & \texttt{redsox}   & : & \texttt{baseball}      & :: & \texttt{BostonBruins}  & : & \texttt{hockey}          \\
\texttt{boston}  & + & \texttt{baseball} & = & \texttt{redsox}          & \texttt{redsox}   & : & \texttt{baseball}      & :: & \texttt{Patriots}      & : & \texttt{nfl}             \\ \cline{6-12}
\texttt{boston}  & + & \texttt{hockey}   & = & \texttt{BostonBruins}    &  \texttt{\textbf{university}} & : & \texttt{\textbf{city}}            & :: &  \texttt{\textbf{university}}      & : & \texttt{\textbf{city}}              \\ \cline{6-12}
\texttt{boston}  & + & \texttt{nba}      & = & \texttt{bostonceltics}   & \texttt{mcgill}   & : & \texttt{montreal}      & :: & \texttt{UBC}           & : & \texttt{vancouver}       \\
\texttt{boston}  & + & \texttt{nfl}      & = & \texttt{Patriots}        & \texttt{mcgill}   & : & \texttt{montreal}      & :: & \texttt{UofT}          & : & \texttt{toronto}
\end{tabularx}
\caption{Examples of subreddit embedding evaluation, based on our composition and analogy tests.}
\label{tab:eval_test}
\end{table*}

In total, we ran 157 composition tests and 6349 analogy tests. In 81\% of cases, the correct answer to a composition test was in the top five most similar subreddits. Similarly, in 84\% of cases, the correct answer to an analogy test was in the top five most similar subreddits. Examples are highlighted in Table \ref{tab:eval_test}, and we note that they are in line with the results reported in the original paper.

\begin{table*}[!t]
\centering
\scriptsize
\begin{tabularx}{\linewidth}{XXXX}
\multicolumn{4}{l}{} \\ \hline 
\texttt{\textbf{gaybros}}      & \texttt{\textbf{Blackfellas}} & \texttt{\textbf{trans}}          & \texttt{\textbf{AgainstHateSubreddits}}    \\ \hline
\texttt{askgaybros}            & \texttt{blackladies}          & \texttt{transpositive}           & \texttt{Fuckthealtright}                    \\
\texttt{gay}                   & \texttt{BlackHair}            & \texttt{ask\_transgender}        & \texttt{TopMindsOfReddit}                   \\
\texttt{gaymers}               & \texttt{racism}               & \texttt{MtF}                     & \texttt{beholdthemasterrace}                \\
\multicolumn{4}{l}{} \\ \hline 
\texttt{\textbf{4chan}}        & \texttt{\textbf{CoonTown}}    & \texttt{\textbf{GenderCritical}} & \texttt{\textbf{MGTOW}}                     \\ \hline
\texttt{ImGoingToHellForThis}  & \texttt{GreatApes}            & \texttt{itsafetish}              & \texttt{WhereAreAllTheGoodMen}              \\
\texttt{classic4chan}          & \texttt{WhiteRights}          & \texttt{GCdebatesQT}             & \texttt{TheRedPill}                         \\
\texttt{CringeAnarchy}         & \texttt{AntiPOZi}             & \texttt{Gender\_Critical}        & \texttt{asktrp}                             \\                        \\
\multicolumn{4}{l}{} \\ \hline 
\texttt{\textbf{changemyview}} & \texttt{\textbf{hiphop}}      & \texttt{\textbf{cars}}           & \texttt{\textbf{relationships}}            \\ \hline
\texttt{PoliticalDiscussion}   & \texttt{90sHipHop}            & \texttt{Autos}                   & \texttt{AskWomen}                           \\
\texttt{bestof}                & \texttt{rap}                  & \texttt{BMW}                     & \texttt{relationship\_advice}               \\
\texttt{TrueAskReddit}         & \texttt{hiphop101}            & \texttt{carporn}                 & \texttt{offmychest}                         \\
\end{tabularx}
\caption{Top three subreddits by cosine similarity to each subreddit in bold (experiments run on top five).}
\label{tab:subreddit_neighbourhood}
\end{table*}

\subsection{Context Insensitive Classifiers}
To assess the importance of community context, we run a series of context sensitive and context insensitive experiments. We run all experiments using a 5-fold cross validation in order to label the entire corpus. Moreover, we use stratified sampling to ensure a uniform distribution of slurs, subreddits, and labels across all folds. Below, we describe the models used for our context insensitive experiments.

\vspace{2mm}

\noindent \textbf{\texttt{(LOG-REG)}}
Our first classifier is a Logistic Regression with L2 regularization. We preprocess the corpus by lowercasing and stemming the text, removing stop words, and masking user mentions and URLs prior to tokenization. Each token is then weighed using \texttt{TF-IDF} to create unigram, bigram, and trigram features. We use \texttt{scikit-learn} to create our classification pipeline.

\vspace{2mm}

\noindent{\textbf{\texttt{(BERT)}}}
Our second classifier is \texttt{BERT}. We use \texttt{BERT-BASE} pre-trained on uncased data with the \texttt{AdamW} optimizer, which has a final linear layer. It takes the top-level embedding of the \texttt{[CLS]} token as input. We do fine-tuning over four epochs with a batch size of 32, and we choose a learning rate of 2e-05 and epsilon 1e-8\footnote{All \texttt{BERT} experiments were performed on Google Colab with Tesla V100-SXM2-16GB GPU, and we use \texttt{BERTForSequenceClassification} from \texttt{huggingface} for our implementation.}.

\vspace{2mm}

\centerline{\texttt{[CLS]} $c$ \texttt{[SEP]}}

\vspace{2mm}

\noindent \textbf{\texttt{(PERSPECTIVE)}} We use a publicly available commercial tool for toxicity detection\footnote{\url{www.perspectiveapi.com}}. It is a CNN-based model that is trained on a high volume of user-generated comments across social media platforms. While the tool is updated by PERSPECTIVE, the API cannot be retrained, fine-tuned, or modified. We use 0.8 as our threshold for \texttt{DEG}.

\subsection{Context Sensitive Classifiers}
Below, we describe the models used for our context sensitive experiments.

\vspace{2mm}

\noindent \textbf{\texttt{(LOG-REG-COMM)}} We use the same setup as in \texttt{LOG-REG}, but we include an additional feature for the name of each subreddit that comments are sourced from. This is done with the purpose of incorporating a social prior with which the algorithm can contextualize the comment text.

\vspace{2mm}

\noindent \textbf{\texttt{(BERT-COMM)}} We concatenate the name of each source subreddit to the beginning of each text before passing the comment to \texttt{BERT}. 

\vspace{2mm}

\centerline{\texttt{[CLS]} $s$ + $c$ \texttt{[SEP]}}

\vspace{2mm}

\noindent{\textbf{\texttt{(BERT-COMM-SEP)}}} In our second variant for context sensitivity, we use the sentence entailment format for \texttt{BERT}. This model concatenates
the comment with the source subreddit, separated by \texttt{BERT}’s \texttt{[SEP]} token. The model is fine-tuned in the same way as our other \texttt{BERT} models. 

\vspace{2mm}

\centerline{\texttt{[CLS]} $c$ \texttt{[SEP]} $s$ \texttt{[SEP]}}

\vspace{2mm}

\begin{table*}[!t]
\centering
\small
\begin{tabularx}{\linewidth}{lYYYY|YYYY}
                       & \multicolumn{4}{c}{\textbf{Performance}}                               & \multicolumn{4}{c}{\textbf{\% Classified} \texttt{DEG}}    \\
                       & \textbf{Accuracy} & \textbf{Precision} & \textbf{Recall} & \textbf{F1} & \texttt{DEG} & \texttt{NDNA} & \texttt{APR} & \texttt{HOM} \\ \hline
\texttt{PERSPECTIVE}   & 0.6132            & 0.6147             & 0.6102          & 0.6079      & 70.75\%      & 53.10\%       & 53.16\%      & 10.71\%      \\
\texttt{LOG-REG}       & 0.8003            & 0.8009             & 0.7994          & 0.7997      & 82.85\%      & 22.46\%       & 61.30\%      & 16.67\%      \\
\texttt{LOG-REG-COMM}  & 0.8002            & 0.8001             & 0.7999          & 0.8000      & 81.10\%      & 20.53\%       & 58.95\%      & 15.67\%      \\
\texttt{BERT}          & 0.8856            & 0.8854             & 0.8857          & 0.8855      & 88.06\%      & 10.26\%       & 47.20\%      & 6.31\%       \\
\texttt{BERT-COMM}     & 0.8905            & 0.8904             & 0.8908          & 0.8905      & 88.08\%      & 9.38\%        & 42.31\%      & 5.36\%       \\
\texttt{BERT-COMM-SEP} & \textbf{0.8930}   & \textbf{0.8930}    & \textbf{0.8934} & \textbf{0.8930} & \textbf{88.12\%}         & 8.95\%        & \textbf{39.60\%}      & 5.11\%       \\
\texttt{BERT-COMM-NGH} & 0.8923            & 0.8924             & 0.8928          & 0.8923      & 87.82\%      & \textbf{8.80\%}        & 39.78\%      & \textbf{4.75\%}      
\end{tabularx}
\caption{Results from our classification task. We report the percentage of each gold label that is classified as \texttt{DEG}. This indicates the percentage of true positives for \texttt{DEG} and the percentage of false positives for the other three labels.}
\label{tab:context-results} 
\end{table*}

\noindent{\textbf{\texttt{(BERT-COMM-NGH)}}} We use our trained \texttt{GloVe} embeddings (see Section \ref{sec:glove}) to obtain the five most similar subreddits to each source subreddits. This allows us to build a direct community neighborhood that we concatenate to the source subreddit. We train this variant of \texttt{BERT} using the same sentence entailment format as was described above.

\vspace{2mm}

\centerline{\texttt{[CLS]} $c$ \texttt{[SEP]} $s_1$ $s_2$ ... $s_6$ \texttt{[SEP]}}

\vspace{2mm}

\section{Results}
\subsection{Subreddits Cluster around Social Polarity}
Prior work has established that communities cluster around topics like music, movies, and sports \cite{martin2017community2vec}. We want to examine how subreddit neighbourhoods behave based on the nature of their support towards marginalized groups. We identify eight supportive and antagonistic subreddits and use our \texttt{GloVe} embeddings to extract the three most similar communities to each of them (see: Table \ref{tab:subreddit_neighbourhood}). We make two main observations.

First, we observe that supportive subreddits are most similar to other supportive subreddits that cater towards the same marginalized community. For instance, the neighbourhood of \texttt{gaybros}, a subreddit built for the LGBTQ+ community, contains other subreddits based on pride and support: \texttt{askgaybros}, \texttt{gay}, and \texttt{gaymers}. A similar trend is observed with the neighbours of \texttt{Blackfellas} and \texttt{trans}.

Second, we see that antagonistic subreddits are most similar to other antagonistic subreddits. \texttt{GenderCritical} is contained in a cluster of anti-trans subreddits, \texttt{MGTOW} is near misogynistic subreddits, and \texttt{CoonTown} is surrounded by racist subreddits. This highlights how polarizing communities tend to cluster around other communities with the same, or similar, polarities. 

Figure \ref{fig:sub-cluster} shows the embeddings of a sample of subreddits from \texttt{Slur-Corpus} plotted in two-dimensions using UMAP. There are independent groups for misogynistic, racist, toxic, anti-hate, black, gay, and trans subreddits.

\subsection{Subreddit Context Reduces False Positives}

\begin{table}[!t]
\centering
\small
\begin{tabularx}{\columnwidth}{rlc|c|cl}
                         &                    &\texttt{BERT} & $\cap$ & \multicolumn{2}{c}{\texttt{BERT-COMM-SEP}} \\ \hline
\textbf{FP} &                          & 765    & 1339   & 480         &                              \\
\textbf{TP}  & \ldelim\{{2}{10pt}[1067] & 587    & 17492  & 599         & \rdelim\}{2}{10pt}[1364]     \\
\textbf{TN}  &                          & 480    & 16696  & 765         &                              \\
\textbf{FN} &                          & 599    & 1853   & 587         &                              \\ \hline
                         & 2.68\%                   & 6.11\% & 93.89\%& 6.11\%      & 3.43\%

\end{tabularx}
\caption{The effect of community context on \texttt{BERT} classification outcomes. The column $\cap$ counts the number of comments with identical labels from \texttt{BERT} and \texttt{BERT-COMM-SEP}, while the columns relating to each classifier only describe comments with different labels. The percentages $2.68\%$ and $3.43\%$ represent the share of true positives and negatives for \texttt{BERT} and \texttt{BERT-COMM-SEP}, respectively.}
\label{tab:result_comaprison} 
\end{table}

We present the results from our classification experiments in Table \ref{tab:context-results}\footnote{We report Macro F1.}. The results will be discussed through two lenses: (1) overall performance; and (2) performance by label.

\texttt{BERT}-based models outperformed classifiers based on Logistic Regression. This is unsurprising, given that Transformers are the current state-of-the-art in NLP. However, \texttt{LOG-REG} achieves nearly 20\% higher accuracy than \texttt{PERSPECTIVE}. While this performance gap is likely the result of the data used to train both models, it is concerning given that the Perspective API is widely used as a tool for toxicity detection with both commercial\footnote{Trusted partners include Reddit, The New York Times, The Financial Times, and the Wall Street Journal.} and academic applications \cite{cuthbertson2019women}.

For both \texttt{BERT} and \texttt{LOG-REG}, the addition of subreddit context reduced the number of false positives across all three non-derogatory labels. Performance on \texttt{DEG} comments remained relatively unchanged. The highest increase in performance was seen with \texttt{BERT-COMM-SEP}, which had each source subreddit concatenated to the comment with a middle \texttt{[SEP]} token. Adding subreddit context led to a significant improvement for appropriative text, across which the false positive rate decreased by almost 8\%. For example, ``\textit{Tr*nny} here, some of us are actually really cool.'' was originally misclassified without community context.

Surprisingly, \texttt{BERT-COMM-NGH}, our model with expanded neighbourhood context, showed little improvement over \texttt{BERT-COMM-SEP}. While the identification of \texttt{NDNA} and \texttt{HOM} improved marginally, the false positive rate for appropriative language increased. One possible explanation is that smaller communities did not have a significant presence in the \texttt{Slur-Corpus} (8\% of all subreddits accounted for 80\% of all comments), and consequently the performance gains associated with comments belonging to these subreddits was marginal. We still believe that neighbourhood context is important for determining the nature of niche communities based on their proximity to larger, established supportive or antagonistic communities. Further analysis of this model is required to understand its full potential.

\subsection{Understanding Context Sensitivity}
We call a comment ``context sensitive'' if the addition of context changed its classification label. \texttt{BERT} and \texttt{BERT-COMM-SEP} have comparable performance on the majority of the corpus: 94\% of comments are context insensitive (see Table \ref{tab:result_comaprison}). However, 1364 of the total classification errors made by \texttt{BERT} were rectified with the inclusion of social context. These classifications represented $>3\%$ of the actual corpus, but 56\% of context-sensitive comments within it. In Table \ref{tab:result_examples}, we present examples of top subreddits for both \textit{true positive} and \textit{true negative} context sensitive comments, along with comments for each. The \textit{true positive} comments largely belonged to antagonistic subreddits, while the \textit{true negative} ones belonged to supportive subreddits. Community context helped \texttt{BERT-COMM-SEP} identify community polarity.

\section{Discussion}
Our analysis points to two key resources that would benefit future abusive language research.

\vspace{2mm}

\noindent \textbf{Subreddit embeddings for community sampling.} Systems for abuse detection should reliably identify different variations of abuse (e.g. sexism, racism, etc.), while still exhibiting sensitivity towards non-derogatory comments (e.g. appropriation, reclamation, etc.). One way to achieve this is to ensure content diversity in training data. \citet{kurrek2020towards} specifically use community sampling to achieve this kind of diversity. The authors collect comments from various Reddit communities, but their work is limited by the absence of resources that identify and consolidate supportive or antagonistic subreddits. Instead, they rely on manual data exploration. There are several issues with this approach. First, knowing which communities to look for (and how to find them) requires a high degree of domain knowledge. Second, manual comment analysis is an expensive task, which makes it difficult to scale or reuse as new communities form. Third, this method is prone to overlooking smaller, niche subreddits that would otherwise have been found using a neighborhood exploration of community clusters. We propose the use of subreddit embeddings in future research to further extend efforts on diverse and representative content collection.

\vspace{2mm}

\noindent \textbf{Community context for protecting productive conversations.} One of our primary research objectives was to ensure that detection frameworks do not mistakenly classify productive conversations as abusive. Community contextualized models, based on Logistic Regression and \texttt{BERT}, better identified non-derogatory comments than their context-insensitive counterparts. Context was found to be particularly helpful for identifying appropriative language, resulting in an 8\% increase in accuracy with the addition of a subreddit name. Appropriation is a tool used by marginalized populations to counteract oppression. When abuse detection frameworks misclassify reclamation, they censor the empowerment tools of the very communities that they are installed to protect. Our analysis of the \texttt{Slur-Corpus} suggests that productive conversations tend to happen in safe and supportive social spaces. It is therefore crucial that these spaces be considered for nuanced classification of abuse.

\begin{table}[!t]
\centering
\small
\begin{tabularx}{\linewidth}{X}
\textbf{\normalsize True Positives} \\ \hline

\\ \texttt{\normalsize CringeAnarchy} \\
I am genuinly surpised at a suicidal \textit{tr*nny} \\ \\ \hline

\\ \texttt{\normalsize 4chan} \\ 
This is basically everyday in Atlanta. It's \textit{n*gger}/\textit{sp*c} central. Give a useful warning next time. \\ \\



\textbf{\normalsize True Negatives} \\ \hline



\\ \texttt{\normalsize BlackPeopleTwitter} \\
Shit Britney rides for us too, idk if you seen when she was about to let the hands fly on some dude for calling her security a \textit{n*gger} \\ \\ \hline

\\ \texttt{\normalsize askgaybros} \\ 
Masc bear here. Twinks are my favorite and \textit{f*ggot} is a pretty funny word :b 
\end{tabularx}
\caption{Top subreddits across comments whose labels were correctly classified with the addition of context.}
\label{tab:result_examples} 
\end{table}

\section{Conclusion and Future Work}
The subjectivity of abuse makes it challenging to annotate and detect reliably. One method for making the problem tractable is to position online conversations within a larger context. This paper was an exploration of one type of contextual information: community identity. We found that the context derived from community identity can help in the collection and classification of abusive language. We therefore believe that community context is integral to all stages of abusive language research. We leave as future work the inclusion of community information in existing, platform-agnostic, ensemble detection frameworks.

\bibliography{anthology,custom}

\begin{thebibliography}{69}
\expandafter\ifx\csname natexlab\endcsname\relax\def\natexlab#1{#1}\fi

\bibitem[{Alonso et~al.(2020)Alonso, Saini, and Kov{\'a}cs}]{alonso2020hate}
Pedro Alonso, Rajkumar Saini, and Gy{\"o}rgy Kov{\'a}cs. 2020.
\newblock Hate speech detection using transformer ensembles on the hasoc
  dataset.
\newblock In \emph{International Conference on Speech and Computer}, pages
  13--21. Springer.

\bibitem[{Balci and Salah(2015)}]{balci2015automatic}
Koray Balci and Albert~Ali Salah. 2015.
\newblock Automatic analysis and identification of verbal aggression and
  abusive behaviors for online social games.
\newblock \emph{Computers in Human Behavior}, 53:517--526.

\bibitem[{Baumgartner et~al.(2020)Baumgartner, Zannettou, Keegan, Squire, and
  Blackburn}]{baumgartner2020pushshift}
Jason Baumgartner, Savvas Zannettou, Brian Keegan, Megan Squire, and Jeremy
  Blackburn. 2020.
\newblock The pushshift reddit dataset.
\newblock In \emph{Proceedings of the International AAAI Conference on Web and
  Social Media}, volume~14, pages 830--839.

\bibitem[{Bender et~al.(2021)Bender, Gebru, McMillan-Major, and
  Shmitchell}]{bender2021dangers}
Emily~M Bender, Timnit Gebru, Angelina McMillan-Major, and Shmargaret
  Shmitchell. 2021.
\newblock On the dangers of stochastic parrots: Can language models be too big?
\newblock In \emph{Proceedings of the 2021 ACM Conference on Fairness,
  Accountability, and Transparency}, pages 610--623.

\bibitem[{Bodapati et~al.(2019)Bodapati, Gella, Bhattacharjee, and
  Al-Onaizan}]{bodapati2019neural}
Sravan Bodapati, Spandana Gella, Kasturi Bhattacharjee, and Yaser Al-Onaizan.
  2019.
\newblock Neural word decomposition models for abusive language detection.
\newblock In \emph{Proceedings of the Third Workshop on Abusive Language
  Online}, pages 135--145.

\bibitem[{Bridges(2017)}]{bridges2017gendering}
Judith Bridges. 2017.
\newblock Gendering metapragmatics in online discourse:``mansplaining man gonna
  mansplain{\ldots}''.
\newblock \emph{Discourse, Context \& Media}, 20:94--102.

\bibitem[{Burnap and Williams(2014)}]{burnap2014hate}
Peter Burnap and Matthew~Leighton Williams. 2014.
\newblock Hate speech, machine classification and statistical modelling of
  information flows on twitter: Interpretation and communication for policy
  decision making.
\newblock \emph{Internet, Policy \& Politics}.

\bibitem[{Cao et~al.(2020)Cao, Lee, and Hoang}]{cao2020deephate}
Rui Cao, Roy Ka-Wei Lee, and Tuan-Anh Hoang. 2020.
\newblock Deephate: Hate speech detection via multi-faceted text
  representations.
\newblock In \emph{12th ACM Conference on Web Science}, pages 11--20.

\bibitem[{Chakrabarty et~al.(2019)Chakrabarty, Gupta, and
  Muresan}]{chakrabarty2019pay}
Tuhin Chakrabarty, Kilol Gupta, and Smaranda Muresan. 2019.
\newblock Pay ``attention'' to your context when classifying abusive language.
\newblock In \emph{Proceedings of the Third Workshop on Abusive Language
  Online}, pages 70--79.

\bibitem[{Chatzakou et~al.(2017)Chatzakou, Kourtellis, Blackburn,
  De~Cristofaro, Stringhini, and Vakali}]{chat2017mean}
Despoina Chatzakou, Nicolas Kourtellis, Jeremy Blackburn, Emiliano
  De~Cristofaro, Gianluca Stringhini, and Athena Vakali. 2017.
\newblock Mean birds: Detecting aggression and bullying on twitter.
\newblock In \emph{Proceedings of the 2017 ACM on web science conference},
  pages 13--22.

\bibitem[{Chiril et~al.(2020)Chiril, Moriceau, Benamara, Mari, Origgi, and
  Coulomb-Gully}]{chiril2020he}
Patricia Chiril, V{\'e}ronique Moriceau, Farah Benamara, Alda Mari, Gloria
  Origgi, and Marl{\`e}ne Coulomb-Gully. 2020.
\newblock He said ``who's gonna take care of your children when you are at
  acl?'': Reported sexist acts are not sexist.
\newblock In \emph{Proceedings of the 58th Annual Meeting of the Association
  for Computational Linguistics}, pages 4055--4066.

\bibitem[{Cuthbertson et~al.(2019)Cuthbertson, Kearney, Dawson, Zawaduk,
  Cuthbertson, Gordon-Tighe, and Mathewson}]{cuthbertson2019women}
Lana Cuthbertson, Alex Kearney, Riley Dawson, Ashia Zawaduk, Eve Cuthbertson,
  Ann Gordon-Tighe, and Kory~Wallace Mathewson. 2019.
\newblock Women, politics and twitter: Using machine learning to change the
  discourse.
\newblock In \emph{Proceedings AI for Social Good workshop at NeurIPS}.

\bibitem[{Dadvar et~al.(2013)Dadvar, Trieschnigg, Ordelman, and
  de~Jong}]{dadvar2013improving}
Maral Dadvar, Dolf Trieschnigg, Roeland Ordelman, and Franciska de~Jong. 2013.
\newblock Improving cyberbullying detection with user context.
\newblock In \emph{Proceedings of the 35th European conference on Advances in
  Information Retrieval}, pages 693--696.

\bibitem[{Davidson et~al.(2020)Davidson, Sun, and
  Wojcieszak}]{davidson2020developing}
Sam Davidson, Qiusi Sun, and Magdalena Wojcieszak. 2020.
\newblock Developing a new classifier for automated identification of
  incivility in social media.
\newblock In \emph{Proceedings of the Fourth Workshop on Online Abuse and
  Harms}, pages 95--101.

\bibitem[{Davidson et~al.(2019)Davidson, Bhattacharya, and
  Weber}]{davidson2019racial}
Thomas Davidson, Debasmita Bhattacharya, and Ingmar Weber. 2019.
\newblock Racial bias in hate speech and abusive language detection datasets.
\newblock In \emph{Proceedings of the Third Workshop on Abusive Language
  Online}, pages 25--35. Association for Computational Linguistics.

\bibitem[{Davidson et~al.(2017)Davidson, Warmsley, Macy, and
  Weber}]{davidson2017automated}
Thomas Davidson, Dana Warmsley, Michael Macy, and Ingmar Weber. 2017.
\newblock Automated hate speech detection and the problem of offensive
  language.
\newblock In \emph{Proceedings of the International AAAI Conference on Web and
  Social Media}, volume~11.

\bibitem[{Deal et~al.(2020)Deal, Martinez, Spitzberg, and
  Tsou}]{deal2020definitely}
Bonnie-Elene Deal, Lourdes~S Martinez, Brian~H Spitzberg, and Ming-Hsiang Tsou.
  2020.
\newblock ``i definitely did not report it when i was raped...\#
  webelievechristine\# metoo'': A content analysis of disclosures of sexual
  assault on twitter.
\newblock \emph{Social Media+ Society}, 6.

\bibitem[{Dehghani et~al.(2016)Dehghani, Johnson, Hoover, Sagi, Garten, Parmar,
  Vaisey, Iliev, and Graham}]{dehghani2016purity}
Morteza Dehghani, Kate Johnson, Joe Hoover, Eyal Sagi, Justin Garten,
  Niki~Jitendra Parmar, Stephen Vaisey, Rumen Iliev, and Jesse Graham. 2016.
\newblock Purity homophily in social networks.
\newblock \emph{Journal of Experimental Psychology: General}, 145.

\bibitem[{Del~Vigna et~al.(2017)Del~Vigna, Cimino, Dell'Orletta, Petrocchi, and
  Tesconi}]{del2017hate}
Fabio Del~Vigna, Andrea Cimino, Felice Dell'Orletta, Marinella Petrocchi, and
  Maurizio Tesconi. 2017.
\newblock Hate me, hate me not: Hate speech detection on facebook.
\newblock In \emph{Proceedings of the First Italian Conference on Cybersecurity
  (ITASEC)}, pages 86--95.

\bibitem[{Dixon et~al.(2018)Dixon, Li, Sorensen, Thain, and
  Vasserman}]{dixon2018measuring}
Lucas Dixon, John Li, Jeffrey Sorensen, Nithum Thain, and Lucy Vasserman. 2018.
\newblock Measuring and mitigating unintended bias in text classification.
\newblock In \emph{Proceedings of the 2018 AAAI/ACM Conference on AI, Ethics,
  and Society}, pages 67--73.

\bibitem[{Fan et~al.(2020)Fan, Yu, and Yin}]{fan2020stigmatization}
Lizhou Fan, Huizi Yu, and Zhanyuan Yin. 2020.
\newblock Stigmatization in social media: Documenting and analyzing hate speech
  for covid-19 on twitter.
\newblock \emph{Proceedings of the Association for Information Science and
  Technology}, 57(1):e313.

\bibitem[{Founta et~al.(2019)Founta, Chatzakou, Kourtellis, Blackburn, Vakali,
  and Leontiadis}]{founta2019unified}
Antigoni~Maria Founta, Despoina Chatzakou, Nicolas Kourtellis, Jeremy
  Blackburn, Athena Vakali, and Ilias Leontiadis. 2019.
\newblock A unified deep learning architecture for abuse detection.
\newblock In \emph{Proceedings of the 10th ACM conference on web science},
  pages 105--114.

\bibitem[{Gal{\'a}n-Garc{\'\i}a et~al.(2016)Gal{\'a}n-Garc{\'\i}a, de~la
  Puerta, G{\'o}mez, Santos, and Bringas}]{galan2016supervised}
Patxi Gal{\'a}n-Garc{\'\i}a, Jos{\'e}~Gaviria de~la Puerta, Carlos~Laorden
  G{\'o}mez, Igor Santos, and Pablo~Garc{\'\i}a Bringas. 2016.
\newblock Supervised machine learning for the detection of troll profiles in
  twitter social network: application to a real case of cyberbullying.
\newblock \emph{Logic Journal of the IGPL}, 24(1):42--53.

\bibitem[{Gao and Huang(2017)}]{gao2017detecting}
Lei Gao and Ruihong Huang. 2017.
\newblock Detecting online hate speech using context aware models.
\newblock In \emph{Proceedings of the International Conference Recent Advances
  in Natural Language Processing, {RANLP} 2017}, pages 260--266. INCOMA Ltd.

\bibitem[{Garrick(2006)}]{garrick2006humor}
Jacqueline Garrick. 2006.
\newblock The humor of trauma survivors: Its application in a therapeutic
  milieu.
\newblock \emph{Journal of aggression, maltreatment \& trauma},
  12(1-2):169--182.

\bibitem[{Ging(2019)}]{ging2019alphas}
Debbie Ging. 2019.
\newblock Alphas, betas, and incels: Theorizing the masculinities of the
  manosphere.
\newblock \emph{Men and Masculinities}, 22(4):638--657.

\bibitem[{Greevy(2004)}]{greevy2004automatic}
Edel Greevy. 2004.
\newblock \emph{Automatic text categorisation of racist webpages}.
\newblock Ph.D. thesis, Dublin City University.

\bibitem[{Hom(2008)}]{hom2008semantics}
Christopher Hom. 2008.
\newblock The semantics of racial epithets.
\newblock \emph{The Journal of Philosophy}, 105(8):416--440.

\bibitem[{Janchevski and Gievska(2019)}]{janchevski2019study}
Andrej Janchevski and Sonja Gievska. 2019.
\newblock A study of different models for subreddit recommendation based on
  user-community interaction.
\newblock In \emph{International Conference on ICT Innovations}, pages 96--108.
  Springer.

\bibitem[{Joksimovic et~al.(2019)Joksimovic, Baker, Ocumpaugh, Andres, Tot,
  Wang, and Dawson}]{joksimovic2019automated}
Srecko Joksimovic, Ryan~S Baker, Jaclyn Ocumpaugh, Juan Miguel~L Andres, Ivan
  Tot, Elle~Yuan Wang, and Shane Dawson. 2019.
\newblock Automated identification of verbally abusive behaviors in online
  discussions.
\newblock In \emph{Proceedings of the Third Workshop on Abusive Language
  Online}, pages 36--45.

\bibitem[{Karan and {\v{S}}najder(2019)}]{karan2019preemptive}
Mladen Karan and Jan {\v{S}}najder. 2019.
\newblock Preemptive toxic language detection in wikipedia comments using
  thread-level context.
\newblock In \emph{Proceedings of the Third Workshop on Abusive Language
  Online}, pages 129--134.

\bibitem[{Kennedy et~al.(2020)Kennedy, Jin, Mostafazadeh~Davani, Dehghani, and
  Ren}]{kennedy2020contextualizing}
Brendan Kennedy, Xisen Jin, Aida Mostafazadeh~Davani, Morteza Dehghani, and
  Xiang Ren. 2020.
\newblock Contextualizing hate speech classifiers with post-hoc explanation.
\newblock In \emph{Proceedings of the Annual Meeting of the Association for
  Computational Linguistics}.

\bibitem[{Koufakou et~al.(2020)Koufakou, Pamungkas, Basile, and
  Patti}]{koufakou-etal-2020-hurtbert}
Anna Koufakou, Endang~Wahyu Pamungkas, Valerio Basile, and Viviana Patti. 2020.
\newblock {H}urt{BERT}: Incorporating lexical features with {BERT} for the
  detection of abusive language.
\newblock In \emph{Proceedings of the Fourth Workshop on Online Abuse and
  Harms}, pages 34--43. Association for Computational Linguistics.

\bibitem[{Kumar et~al.(2018)Kumar, Hamilton, Leskovec, and
  Jurafsky}]{kumar2018community}
Srijan Kumar, William~L Hamilton, Jure Leskovec, and Dan Jurafsky. 2018.
\newblock Community interaction and conflict on the web.
\newblock In \emph{Proceedings of the 2018 world wide web conference}, pages
  933--943.

\bibitem[{Kurrek et~al.(2020)Kurrek, Saleem, and Ruths}]{kurrek2020towards}
Jana Kurrek, Haji~Mohammad Saleem, and Derek Ruths. 2020.
\newblock Towards a comprehensive taxonomy and large-scale annotated corpus for
  online slur usage.
\newblock In \emph{Proceedings of the Fourth Workshop on Online Abuse and
  Harms}, pages 138--149.

\bibitem[{Liu and Forss(2014)}]{liu2014combining}
Shuhua Liu and Thomas Forss. 2014.
\newblock Combining n-gram based similarity analysis with sentiment analysis in
  web content classification.
\newblock In \emph{Proceedings of the International Joint Conference on
  Knowledge Discovery, Knowledge Engineering and Knowledge Management-Volume
  1}, pages 530--537.

\bibitem[{Markov and Daelemans(2021)}]{markov2021improving}
Ilia Markov and Walter Daelemans. 2021.
\newblock Improving cross-domain hate speech detection by reducing the false
  positive rate.
\newblock In \emph{Proceedings of the Fourth Workshop on NLP for Internet
  Freedom: Censorship, Disinformation, and Propaganda}.

\bibitem[{Martin(2017)}]{martin2017community2vec}
Trevor Martin. 2017.
\newblock community2vec: Vector representations of online communities encode
  semantic relationships.
\newblock In \emph{Proceedings of the Second Workshop on NLP and Computational
  Social Science}, pages 27--31.

\bibitem[{Massanari(2017)}]{massanari2017gamergate}
Adrienne Massanari. 2017.
\newblock \# gamergate and the fappening: How reddit's algorithm, governance,
  and culture support toxic technocultures.
\newblock \emph{New media \& society}, 19(3):329--346.

\bibitem[{Mathur et~al.(2018)Mathur, Shah, Sawhney, and
  Mahata}]{mathur2018detecting}
Puneet Mathur, Rajiv Shah, Ramit Sawhney, and Debanjan Mahata. 2018.
\newblock Detecting offensive tweets in hindi-english code-switched language.
\newblock In \emph{Proceedings of the Sixth International Workshop on Natural
  Language Processing for Social Media}, pages 18--26.

\bibitem[{McLamore and Ulu{\u{g}}(2020)}]{mclamore2020social}
Quinnehtukqut McLamore and {\"O}zden~Melis Ulu{\u{g}}. 2020.
\newblock Social representations of sociopolitical groups on r/the\_donald and
  emergent conflict narratives: A qualitative content analysis.
\newblock \emph{Analyses of Social Issues and Public Policy}.

\bibitem[{Meyer and Gamb{\"a}ck(2019)}]{meyer2019platform}
Johannes~Skjeggestad Meyer and Bj{\"o}rn Gamb{\"a}ck. 2019.
\newblock A platform agnostic dual-strand hate speech detector.
\newblock In \emph{ACL 2019 The Third Workshop on Abusive Language Online
  Proceedings of the Workshop}. Association for Computational Linguistics.

\bibitem[{Mir{\'o}-Llinares et~al.(2018)Mir{\'o}-Llinares, Moneva, and
  Esteve}]{miro2018hate}
Fernando Mir{\'o}-Llinares, Asier Moneva, and Miriam Esteve. 2018.
\newblock Hate is in the air! but where? introducing an algorithm to detect
  hate speech in digital microenvironments.
\newblock \emph{Crime Science}, 7(1):1--12.

\bibitem[{Mishra et~al.(2018)Mishra, Del~Tredici, Yannakoudakis, and
  Shutova}]{mishra2018author}
Pushkar Mishra, Marco Del~Tredici, Helen Yannakoudakis, and Ekaterina Shutova.
  2018.
\newblock Author profiling for abuse detection.
\newblock In \emph{Proceedings of the 27th international conference on
  computational linguistics}, pages 1088--1098.

\bibitem[{Modha et~al.(2018)Modha, Majumder, and Mandl}]{modha2018filtering}
Sandip Modha, Prasenjit Majumder, and Thomas Mandl. 2018.
\newblock Filtering aggression from the multilingual social media feed.
\newblock In \emph{Proceedings of the First Workshop on Trolling, Aggression
  and Cyberbullying (TRAC-2018)}, pages 199--207.

\bibitem[{Narang and Brew(2020)}]{narang2020abusive}
Kanika Narang and Chris Brew. 2020.
\newblock Abusive language detection using syntactic dependency graphs.
\newblock In \emph{Proceedings of the Fourth Workshop on Online Abuse and
  Harms}, pages 44--53.

\bibitem[{Nobata et~al.(2016)Nobata, Tetreault, Thomas, Mehdad, and
  Chang}]{nobata2016abusive}
Chikashi Nobata, Joel Tetreault, Achint Thomas, Yashar Mehdad, and Yi~Chang.
  2016.
\newblock Abusive language detection in online user content.
\newblock In \emph{Proceedings of the 25th international conference on world
  wide web}, pages 145--153.

\bibitem[{Oliva et~al.(2021)Oliva, Antonialli, and Gomes}]{oliva2021fighting}
Thiago~Dias Oliva, Dennys~Marcelo Antonialli, and Alessandra Gomes. 2021.
\newblock Fighting hate speech, silencing drag queens? artificial intelligence
  in content moderation and risks to lgbtq voices online.
\newblock \emph{Sexuality \& Culture}, 25(2):700--732.

\bibitem[{Papegnies et~al.(2017)Papegnies, Labatut, Dufour, and
  Linares}]{papegnies2017graph}
Etienne Papegnies, Vincent Labatut, Richard Dufour, and Georges Linares. 2017.
\newblock Graph-based features for automatic online abuse detection.
\newblock In \emph{International conference on statistical language and speech
  processing}, pages 70--81. Springer.

\bibitem[{Pavlopoulos et~al.(2020)Pavlopoulos, Sorensen, Dixon, Thain, and
  Androutsopoulos}]{pavlopoulos2020toxicity}
John Pavlopoulos, Jeffrey Sorensen, Lucas Dixon, Nithum Thain, and Ion
  Androutsopoulos. 2020.
\newblock Toxicity detection: Does context really matter?
\newblock In \emph{Proceedings of the 58th Annual Meeting of the Association
  for Computational Linguistics}.

\bibitem[{Qian et~al.(2018)Qian, ElSherief, Belding, and
  Wang}]{qian2018leveraging}
Jing Qian, Mai ElSherief, Elizabeth Belding, and William~Yang Wang. 2018.
\newblock Leveraging intra-user and inter-user representation learning for
  automated hate speech detection.
\newblock In \emph{Proceedings of the 2018 Conference of the North American
  Chapter of the Association for Computational Linguistics: Human Language
  Technologies}, volume~2, pages 118--123.

\bibitem[{Sap et~al.(2019)Sap, Card, Gabriel, Choi, and Smith}]{sap2019risk}
Maarten Sap, Dallas Card, Saadia Gabriel, Yejin Choi, and Noah~A Smith. 2019.
\newblock The risk of racial bias in hate speech detection.
\newblock In \emph{Proceedings of the 57th Annual Meeting of the Association
  for Computational Linguistics}, pages 1668--1678.

\bibitem[{Saveski et~al.(2021)Saveski, Roy, and Roy}]{saveski2021structure}
Martin Saveski, Brandon Roy, and Deb Roy. 2021.
\newblock The structure of toxic conversations on twitter.
\newblock In \emph{Proceedings of the Web Conference 2021}, pages 1086--1097.

\bibitem[{Silva et~al.(2017)Silva, Santana, Lobato, and
  Pinheiro}]{silva2017methodology}
Wendel Silva, {\'A}damo Santana, F{\'a}bio Lobato, and M{\'a}rcia Pinheiro.
  2017.
\newblock A methodology for community detection in twitter.
\newblock In \emph{Proceedings of the International Conference on Web
  Intelligence}, pages 1006--1009.

\bibitem[{Soliman et~al.(2019)Soliman, Hafer, and
  Lemmerich}]{soliman2019characterization}
Ahmed Soliman, Jan Hafer, and Florian Lemmerich. 2019.
\newblock A characterization of political communities on reddit.
\newblock In \emph{Proceedings of the 30th ACM conference on hypertext and
  Social Media}, pages 259--263.

\bibitem[{Stewart and Spiro(2021)}]{stewart2021manosphere}
Leo~G. Stewart and Emma~S. Spiro. 2021.
\newblock Nobody puts redditor in a binary: Digital demography, collective
  identities, and gender in a subreddit network.
\newblock In \emph{Proceedings of the 24th ACM Conference on Computer-Supported
  Cooperative Work and Social Computing}. Association for Computing Machinery.

\bibitem[{Tulkens et~al.(2016)Tulkens, Hilte, Lodewyckx, Verhoeven, and
  Daelemans}]{tulkens2016dictionary}
St{\'e}phan Tulkens, Lisa Hilte, Elise Lodewyckx, Ben Verhoeven, and Walter
  Daelemans. 2016.
\newblock A dictionary-based approach to racism detection in dutch social
  media.
\newblock In \emph{Proceedings of the First Workshop on Text Analytics for
  Cybersecurity and Online Safety}, page~11. LREC.

\bibitem[{Unsv{\aa}g and Gamb{\"a}ck(2018)}]{unsvaag2018effects}
Elise~Fehn Unsv{\aa}g and Bj{\"o}rn Gamb{\"a}ck. 2018.
\newblock The effects of user features on twitter hate speech detection.
\newblock In \emph{Proceedings of the 2nd workshop on abusive language online
  (ALW2)}, pages 75--85.

\bibitem[{Van~Hee et~al.(2018)Van~Hee, Jacobs, Emmery, Desmet, Lefever,
  Verhoeven, De~Pauw, Daelemans, and Hoste}]{van2018automatic}
Cynthia Van~Hee, Gilles Jacobs, Chris Emmery, Bart Desmet, Els Lefever, Ben
  Verhoeven, Guy De~Pauw, Walter Daelemans, and V{\'e}ronique Hoste. 2018.
\newblock Automatic detection of cyberbullying in social media text.
\newblock \emph{PloS one}, 13.

\bibitem[{Vidgen et~al.(2019)Vidgen, Harris, Nguyen, Tromble, Hale, and
  Margetts}]{vidgen2019challenges}
Bertie Vidgen, Alex Harris, Dong Nguyen, Rebekah Tromble, Scott Hale, and Helen
  Margetts. 2019.
\newblock Challenges and frontiers in abusive content detection.
\newblock In \emph{Proceedings of the Third Workshop on Abusive Language
  Online}, pages 80--93.

\bibitem[{Waller and Anderson(2019)}]{waller2019generalists}
Isaac Waller and Ashton Anderson. 2019.
\newblock Generalists and specialists: Using community embeddings to quantify
  activity diversity in online platforms.
\newblock In \emph{The World Wide Web Conference}, pages 1954--1964.

\bibitem[{Warner and Hirschberg(2012)}]{warner2012detecting}
William Warner and Julia Hirschberg. 2012.
\newblock Detecting hate speech on the world wide web.
\newblock In \emph{Proceedings of the second workshop on language in social
  media}, pages 19--26.

\bibitem[{Waseem and Hovy(2016)}]{waseem2016hateful}
Zeerak Waseem and Dirk Hovy. 2016.
\newblock Hateful symbols or hateful people? predictive features for hate
  speech detection on twitter.
\newblock In \emph{Proceedings of the NAACL student research workshop}, pages
  88--93.

\bibitem[{Wiegand et~al.(2019)Wiegand, Ruppenhofer, and
  Kleinbauer}]{wiegand2019detection}
Michael Wiegand, Josef Ruppenhofer, and Thomas Kleinbauer. 2019.
\newblock Detection of abusive language: the problem of biased datasets.
\newblock In \emph{Proceedings of the 2019 Conference of the North American
  Chapter of the Association for Computational Linguistics: Human Language
  Technologies}, pages 602--608.

\bibitem[{Xia et~al.(2020)Xia, Field, and Tsvetkov}]{xia2020demoting}
Mengzhou Xia, Anjalie Field, and Yulia Tsvetkov. 2020.
\newblock Demoting racial bias in hate speech detection.
\newblock In \emph{Proceedings of the Eighth International Workshop on Natural
  Language Processing for Social Media}, pages 7--14. ACL.

\bibitem[{Zampieri et~al.(2020)Zampieri, Nakov, Rosenthal, Atanasova,
  Karadzhov, Mubarak, Derczynski, Pitenis, and
  {\c{C}}{\"o}ltekin}]{zampieri-etal-2020-semeval}
Marcos Zampieri, Preslav Nakov, Sara Rosenthal, Pepa Atanasova, Georgi
  Karadzhov, Hamdy Mubarak, Leon Derczynski, Zeses Pitenis, and
  {\c{C}}a{\u{g}}r{\i} {\c{C}}{\"o}ltekin. 2020.
\newblock {S}em{E}val-2020 task 12: Multilingual offensive language
  identification in social media ({O}ffens{E}val 2020).
\newblock In \emph{Proceedings of the Fourteenth Workshop on Semantic
  Evaluation}, pages 1425--1447. International Committee for Computational
  Linguistics.

\bibitem[{Zhang et~al.(2020)Zhang, Bai, Zhang, Bai, Zhu, and
  Zhao}]{zhang2020demographics}
Guanhua Zhang, Bing Bai, Junqi Zhang, Kun Bai, Conghui Zhu, and Tiejun Zhao.
  2020.
\newblock Demographics should not be the reason of toxicity: Mitigating
  discrimination in text classifications with instance weighting.
\newblock In \emph{Proceedings of the 58th Annual Meeting of the Association
  for Computational Linguistics}, pages 4134--4145.

\bibitem[{Zhang et~al.(2018)Zhang, Robinson, and Tepper}]{zhang2018detecting}
Ziqi Zhang, David Robinson, and Jonathan Tepper. 2018.
\newblock Detecting hate speech on twitter using a convolution-gru based deep
  neural network.
\newblock In \emph{European semantic web conference}, pages 745--760. Springer.

\bibitem[{Ziems et~al.(2020)Ziems, Vigfusson, and
  Morstatter}]{ziems2020aggressive}
Caleb Ziems, Ymir Vigfusson, and Fred Morstatter. 2020.
\newblock Aggressive, repetitive, intentional, visible, and imbalanced:
  Refining representations for cyberbullying classification.
\newblock In \emph{Proceedings of the International AAAI Conference on Web and
  Social Media}, volume~14, pages 808--819.

\end{thebibliography}
\bibliographystyle{acl_natbib}
\end{document}